\documentclass[conference]{IEEEtran}
\usepackage{times}

\usepackage[numbers]{natbib}
\usepackage{multicol}
\usepackage[bookmarks=true]{hyperref}
\usepackage{graphicx}
\usepackage{amsmath}
\usepackage{booktabs}
\usepackage{multirow}

\begin{document}

\title{Data Analogies Enable \\
Efficient Cross-Embodiment Transfer}

\author{Jonathan Yang, Chelsea Finn, Dorsa Sadigh  \\
Stanford University}



%

\maketitle
\begin{figure*}[t]
\centering
\includegraphics[width=\textwidth]{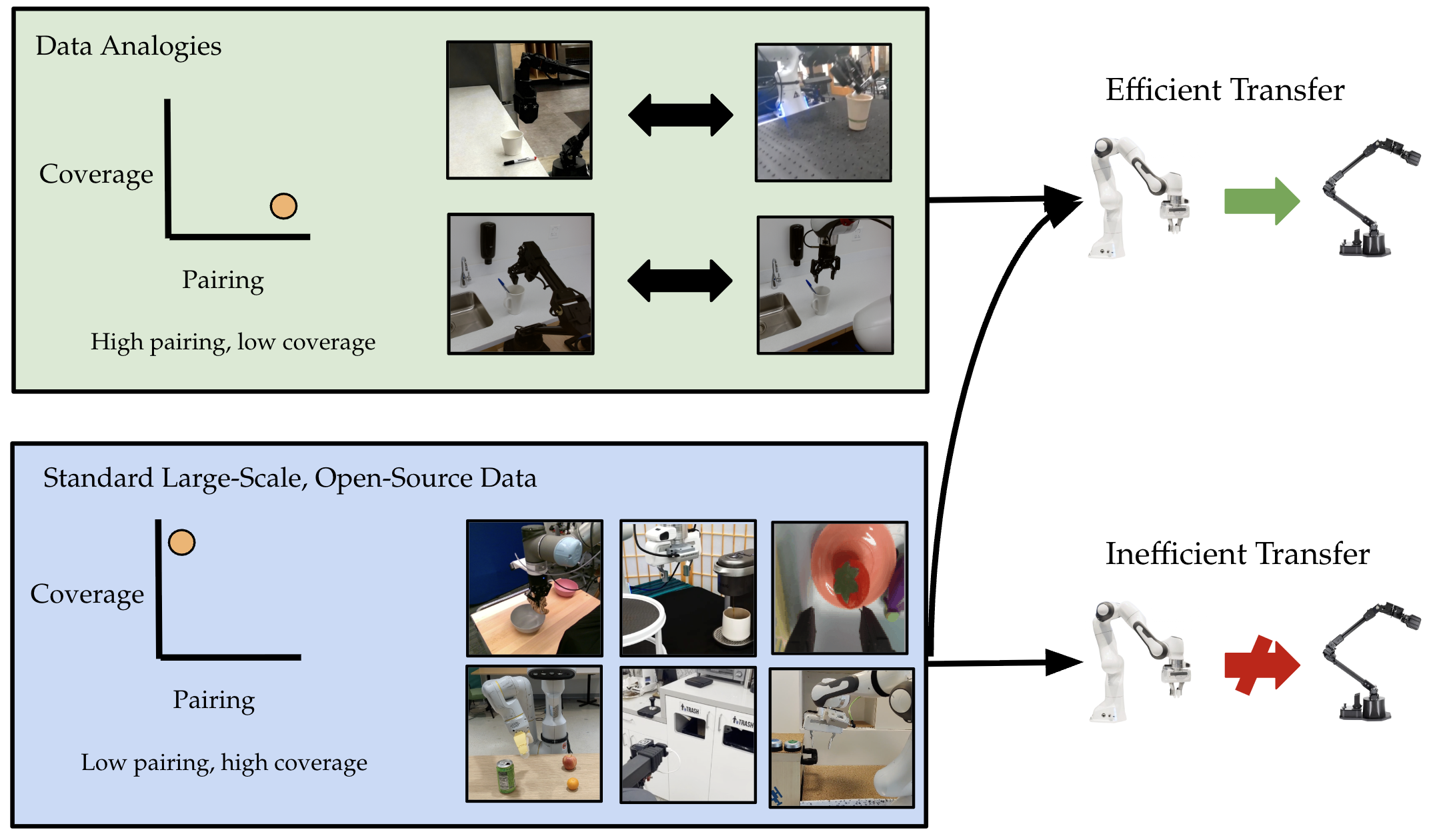}
\caption{\textbf{Cross-Embodiment Data Analogies.} We study how to collect data so that demonstrations from one robot directly help another. Our data-centric recipe composes breadth (\emph{coverage} across viewpoints, morphologies, and scenes). Then, we search different data scaling strategies to find one that leads to high performance under a fixed budget. We find that datasets with high pairing between scenes and tasks as well as high coverage lead to high transfer performance.}
\label{fig:teaser}
\end{figure*}

\begin{abstract}
Generalist robot policies are trained on demonstrations collected across a wide variety of robots, scenes, and viewpoints. Yet it remains unclear how to best organize and scale such heterogeneous data so that it genuinely improves performance in a given target setting. In this work, we ask: what form of demonstration data is most useful for enabling transfer across robot set-ups? We conduct controlled experiments that vary end-effector morphology, robot platform appearance, and camera perspective, and compare the effects of simply scaling the number of demonstrations against systematically broadening the diversity in different ways. Our simulated experiments show that while perceptual shifts such as viewpoint benefit most from broad diversity, morphology shifts benefit far less from unstructured diversity and instead see the largest gains from data analogies, i.e. paired demonstrations that align scenes, tasks, and/or trajectories across different embodiments. Informed by the simulation results, we improve real-world cross-embodiment transfer success by an average of $22.5\%$ over large-scale, unpaired datasets by changing only the composition of the data. For further information, please visit 
\href{https://data-analogies.github.io/}{https://data-analogies.github.io/}.
\end{abstract}

\IEEEpeerreviewmaketitle

\section{Introduction}
\label{sec:intro}
Generalist robot policies are now trained on increasingly large cross-embodiment datasets spanning many robots, morphologies, and viewpoints. At first glance, these results suggest that cross-embodiment learning “just works”—that scaling demonstrations across diverse embodiments naturally enables transfer. Yet the reality is far less clear: we do not know what is actually being transferred when data from other robots is introduced. Current datasets leave critical axes of generalization underrepresented, such as systematic variation in morphology, camera viewpoint, and environment. This raises a key question: are models truly learning useful invariances across morphology and viewpoint, or are their apparent successes merely artifacts of scale? This uncertainty highlights a central gap in the field: we still lack a principled understanding of what kinds of cross-embodiment data actually help a policy adapt to a new robot with only limited data from the target robot setup.

Prior work has largely pursued two strategies for cross-embodiment transfer. Some works simply aggregate demonstrations across robots and environments, banking on diversity at scale to improve robustness \cite{dasari19robonet, o2023open, wang2025unifiedvisionlanguageactionmodel}. While this more implicit approach to transfer produces strong, consistent gains, it makes the magnitude of true transfer hard to diagnose, leaving it unclear whether the policy's performance gain is due to direct motion transfer, high-level behavioral transfer, or merely visual regularization. In contrast, explicit alignment methods, such as generative inpainting, offer a way to directly allow demonstrations from one robot in one scene to be used in another, providing more interpretable assumptions about the mechanism of transfer ~\cite{chen2024miragecrossembodimentzeroshotpolicy, lepert2025shadowleveragingsegmentationmasks, lepert2025masqueradelearninginthewildhuman, pace2025xdiffusiontrainingdiffusionpolicies}. However, these methods are less easily scaled, with many assumptions about the scene that fail to scale across the diversity of morphologies and viewpoints in robotics. In this paper, we aim to achieve the best of both worlds: maintaining the scalability of basic data aggregation while achieving the direct, high-fidelity transfer characteristic of explicit alignment methods.

We study how data collection strategies can affect cross-embodiment transfer, when the target robot has only a small amount of collected data. Our investigation considers basic imitation learning on cross-embodiment data, \emph{without} any special architectural or algorithmic changes such as in-painting. While data diversity is known to contribute to generalization, we also consider an orthogonal aspect of the data, which is the extent to which the data contains analogous examples across robot set-ups. These \emph{data analogies} entail demonstrations from different robot embodiments with similar environments, tasks, and/or trajectories, and have the potential to more explicitly teach the model how different robots relate to each other. See Figure~\ref{fig:teaser} for an example.

The main contribution of this paper is an empirical investigation into how dataset composition impacts cross-embodiment transfer. We focus our experiments on three main aspects of cross-embodiment transfer: camera viewpoint, end-effector morphology, and robot appearance. Our key finding is that data analogies lead to significantly improved transfer compared to simply scaling data diversity, especially for transfer across different morphologies. We observe that our data collection strategy outperforms standard training on large, open-source datasets such as OXE, achieving an average of $19\%$ higher success rate in simulation and $22.5\%$ higher success rate in real-world experiments.

\section{Related Works} \smallskip \noindent The pursuit of generalist robot policies has historically focused on leveraging large, diverse robot datasets~\cite{dasari19robonet, o2023open, intelligence2025pi05visionlanguageactionmodelopenworld}. However, policies trained in this manner often failed to generalize to even simple, low-level variations, such as objects of different colors, revealing a critical lack of fundamental visual capabilities. This limitation highlighted the need for a conceptual "glue" to connect low-level actions with high-level semantic understanding. Vision-Language-Action (VLA) models~\cite{brohan2022rt,  brohan2023rt, kim24openvla, black2024pi_0, szot2024multimodal, team2025gemini, intelligence2025pi, nvidia2025gr00tn1openfoundation} emerged as this glue. By fine-tuning internet-scale vision-language models (VLMs) to predict robot actions, VLAs inherit a rich visual and semantic knowledge of the world, which significantly enhances policy generalization and allows them to ingest large-scale, heterogeneous robot datasets~\cite{kim24openvla, belkhale2024minivla, black2024pi_0}.

The success of VLAs in learning from diverse, data naturally raised a new question: could policies also leverage data from other robots? Initial findings confirmed that large-scale, cross-embodiment datasets could facilitate transfer~\cite{o2023open, doshi2024scalingcrossembodiedlearningpolicy, wang2025unifiedvisionlanguageactionmodel}. However, it was unclear what type of transfer was occurring or if this highly heterogeneous data was being utilized effectively. This ambiguity sparked two main lines of research: (1) methods that explicitly align data across embodiments, and (2) methods that learn common representations to implicitly bridge the embodiment gap. Explicit alignment methods include generative approaches, such as masking~\cite{lepert2025shadowleveragingsegmentationmasks, lepert2025masqueradelearninginthewildhuman, rayyan2025mvumiscalablemultiviewinterface} and inpainting robot-specific features~\cite{chen2024roviaugrobotviewpointaugmentation, chen2024miragecrossembodimentzeroshotpolicy, pace2025xdiffusiontrainingdiffusionpolicies}, as well as motion retargeting~\cite{Aberman_2020, choi2021selfsupervisedmotionretargetingsafety, yan2024imitationnetunsupervisedhumantorobotmotion, cao2025gdreamgraphconditioneddiffusionretargeting, allu2025hrt1oneshothumantorobottrajectory}. A key drawback of these generative methods is their reliance on assumptions that may not scale, such as the ability to perfectly mask and inpaint novel robot embodiments in arbitrary scenes. Meanwhile, implicit methods focus on learning common representations, often by training large-scale models to project diverse robot data into a shared latent space~\cite{yang2023polybot, yang2024pushinglimitscrossembodimentlearning, Doshi24-crossformer, pace2025xdiffusiontrainingdiffusionpolicies, wang2025unifiedvisionlanguageactionmodel, liu2025immimiccrossdomainimitationhuman}. These methods offer greater scalability, as they do not require embodiment-specific modules for data translation.

While these model-centric approaches to alignment and representation have shown promise, they largely treat the underlying data distribution as a given. This overlooks a crucial, emerging direction: focusing on the data distribution itself. For example, recent works have found that carefully collecting and rebalancing data distributions for training VLAs can unlock gains in generalization and transfer~\cite{gao2024efficientdatacollectionrobotic, hejna2024remixoptimizingdatamixtures, xing2025shortcutlearninggeneralistrobot, shi2025diversityneedscalablerobotic, hu2025datascalinglawsimitation}. We argue that this data-centric principle is particularly crucial for cross-embodiment learning. Prior works have seen some success in improving robot generalization by expanding the data coverage or diversity of demonstration data  \cite{11127989, liu2025egozerorobotlearningsmart, bi2025hrdthumanmanipulationenhanced, yuan2025motiontranshumanvrdata}. However, it is still unclear what methods for scaling are useful to cross the embodiment gap. Analogous to how VLAs provided a high-level glue for generalization, a new data-centric glue is needed to bridge the gaps in embodiment. This work investigates how to scale embodiments by prioritizing the right axes of data coverage and data distribution for high-fidelity cross-embodiment transfer.

\section{Cross-Embodiment Data Analogies}
We consider the problem of transferring behaviors from one or multiple robot embodiments to another embodiment. Zero-shot transfer without any data from the target embodiment is impractically difficult, so we study the few-shot adaptation problem where we have access to a small number of demonstrations from the target embodiment, but where these demonstrations are not directly for the target task or scenario (e.g. with different objects, skills, or environments than those evaluated on). 
In this paper, we study how robot data should be collected and organized to enable this form of cross-embodiment generalization. In particular, we find that \emph{data analogies}, or loosely paired demonstrations across embodiments that preserve task-relevant structure despite domain shifts such as viewpoint, end-effector morphology, and appearance,  play a central role in enabling effective few-shot cross-embodiment adaptation.

Concretely, we assume a cross-embodiment dataset $\mathcal{D}=\{(e,\tau)\}$, where $e\in\mathcal{E}$ indexes the embodiment (platform, end-effector, viewpoint) and each trajectory $\tau$ is a sequence $\big((o_1^e,a_1^e),\ldots,(o_T^e,a_T^e)\big)$. Let $e^\star$ denote a target robot with a small few-shot dataset $\mathcal{D}_{e^\star}^{\text{few}}$. We start from a pretrained base policy $\pi_{0.5}$ trained on large-scale, generic robot data. Our goal is to use the cross-embodiment dataset $\mathcal{D}$ as a \emph{bridging signal} that exposes the policy to structured variation across embodiments, enabling it to reuse task-relevant structure when adapting to the target robot. Due to computational constraints, we introduce the cross-embodiment dataset $\mathcal{D}$ only during fine-tuning, alongside $\mathcal{D}_{e^\star}^{\text{few}}$, to adapt the policy to the target embodiment. Our objective is for the resulting policy $\pi_\theta$ to effectively leverage demonstrations from $e \neq e^\star$ to perform the same task on $e^\star$ preserving task-relevant structure while adapting embodiment-specific control—under a fixed $\mathcal{D}_{e^\star}^{\text{few}}$ budget. Our aim is to identify which data collection strategies enable this form of few-shot cross-embodiment adaptation.

\begin{figure}[t]
\centering
\includegraphics[width=\columnwidth]{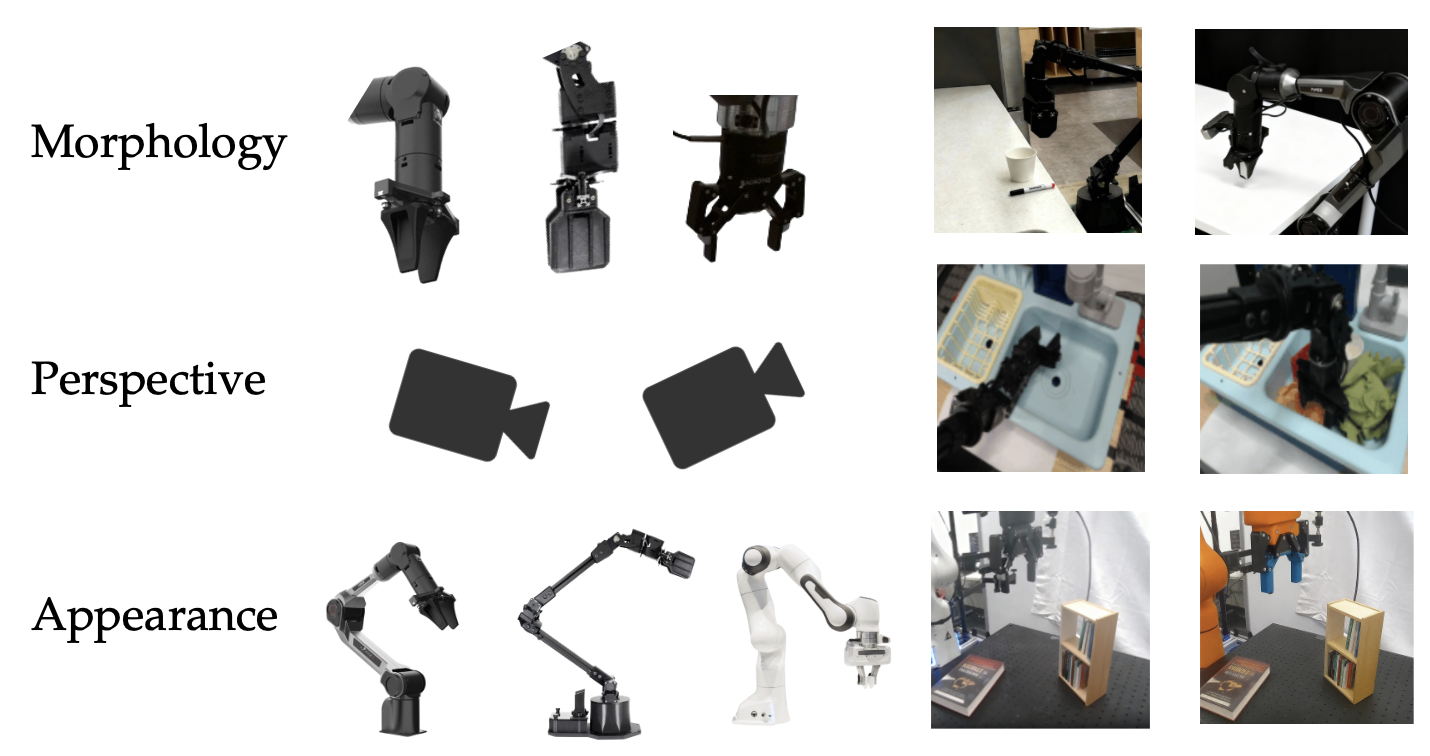}
\caption{\textbf{Domain Shift Axes.} We study the role of data diversity and pairing across three domain shift axes: end-effector morphology, camera perspective, and visual appearance.}
\label{fig:cov_vs_pairing}
\end{figure}

How should we design a data collection strategy that enables effective few-shot cross-embodiment adaptation? We begin by decomposing what a policy must \emph{understand} in order to reuse demonstrations across embodiments. For single-arm manipulators, the cross-embodiment gap is driven by three primary domain shifts: \textbf{viewpoint} (camera pose and intrinsics), \textbf{end-effector morphology} (gripper geometry and arm kinematics), and \textbf{appearance} (textures, lighting, and background). For \emph{each} domain shift, we systematically sweep two orthogonal data collection axes under matched data budgets:

\begin{figure*}[t]
\centering
\includegraphics[width=\textwidth]{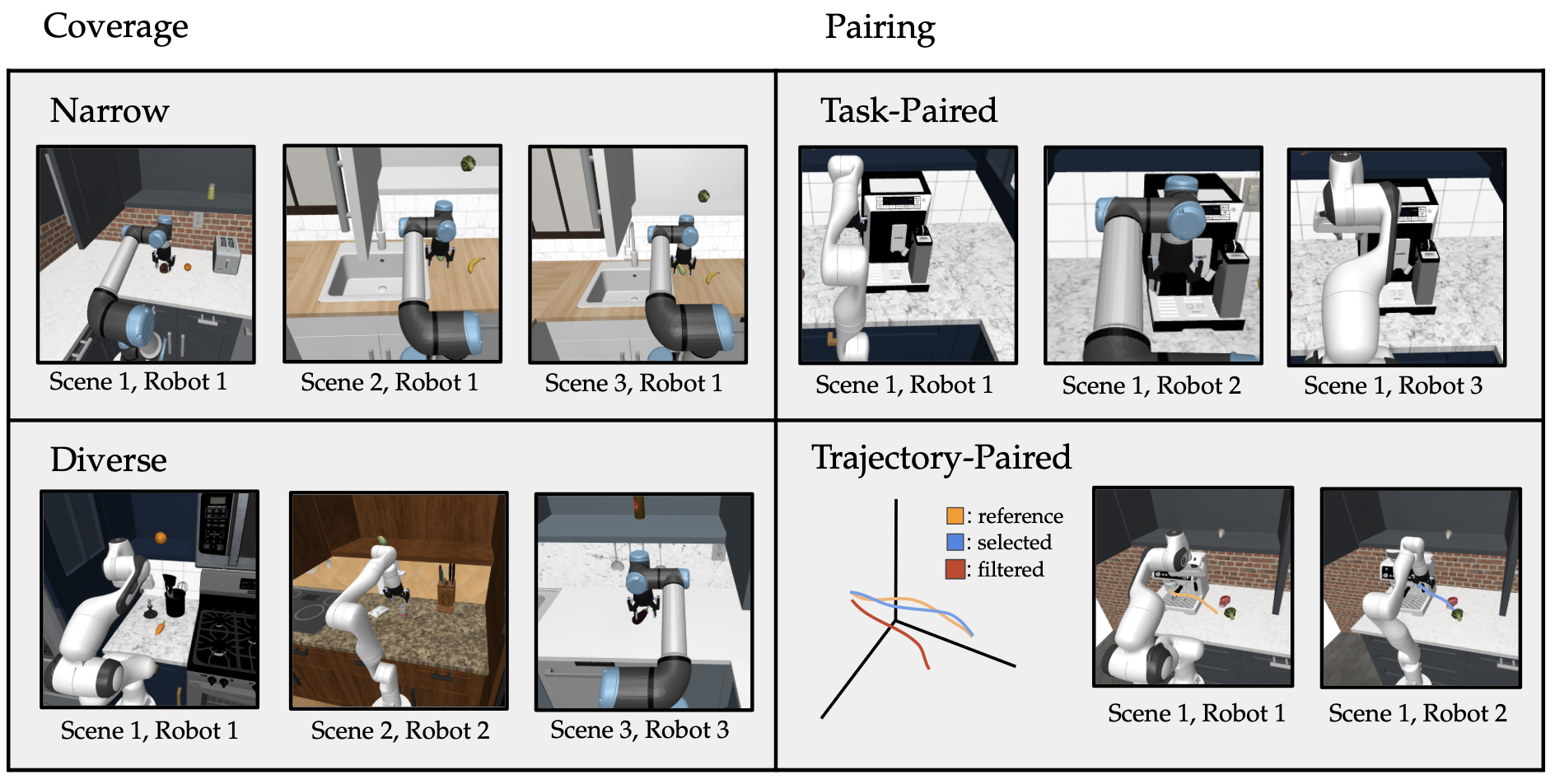}
\caption{\textbf{Coverage versus Pairing.} Simulation images depicting the data collection strategies. Coverage is the diversity of data on the generalization axis, while pairing is the similarity of the tasks or trajectories in the data.}
\label{fig:cov_vs_pairing}
\end{figure*}

\begin{itemize}
    \item \textbf{Coverage strategy: Targeted vs.\ Diverse.}
    \begin{itemize}
        \item \emph{Targeted} — Select demonstrations that \emph{close gaps relative to the target robot}, using simple coverage criteria: e.g., fill missing bins in camera extrinsics/intrinsics for \textit{viewpoint}, cover specific gripper types/kinematic regimes for \textit{morphology}, or match scene materials/lighting regimes for \textit{appearance}.
        \item \emph{Diverse} — Collect broadly varied demonstrations \emph{without target-aware coverage}, sampling viewpoints, morphologies, or appearances uniformly/randomly across what is available.
    \end{itemize}

    \item \textbf{Cross-robot pairing: Unpaired vs.\ Task-Paired vs.\ Trajectory-Paired.}
    \begin{itemize}
        \item \emph{Unpaired} — Source and target demonstrations are independent; no cross-robot alignment beyond task labels.
        \item \emph{Task-Paired} — Demos correspond to the \emph{same task instance} across robots (same objects/initial conditions/goals), but are only \emph{weakly aligned} (e.g., matched hand-designed keypoints).
        \item \emph{Trajectory-Paired} — A deliberate data collection strategy to capture the \emph{same execution strategy} across embodiments.
        \begin{itemize}
            \item \textbf{In simulation,} this can be achieved by filtering for demonstrations with highly similar object-centric trajectories via dynamic time-warping (DTW). To make this computational alignment scalable for both real and simulated data, trajectories are first downsampled to a fixed length (50 timesteps).
            \item \textbf{In the real world,} this is achieved by collecting demonstrations of the \emph{same task instance} (same scene, objects, and goal) from two different robots, and then computationally aligning the resulting trajectories via DTW. 
        \end{itemize}
    \end{itemize}
\end{itemize}

We aggregate the per-axis strategy datasets into a single \emph{compositional dataset} $\mathcal{D}_{\text{comp}}$, with an equal data budget allocated to each coverage–pairing combination. Across experiments, we find that the most effective data for few-shot cross-embodiment adaptation takes the form of \emph{data analogies}: demonstrations that combine broad coverage across embodiments with strong, trajectory-level pairing that preserves task-relevant structure over time. These paired analogies enable the policy to meaningfully reuse cross-robot demonstrations, rather than treating embodiment variation as unstructured noise.

\section{Experiments}

Our goal is to evaluate \emph{cross-embodiment translation}: can demonstrations collected on one robot help a different robot perform the \emph{same} task with only a few target examples? We align our study around four questions that map directly to the four main figures.

\begin{enumerate}
    \item \textbf{Q1 (Fig.~1): Under a fixed budget, what data collection strategy leads to the highest performance for each generalization axis?}
    \item \textbf{Q2 (Fig.~2): How does our compositional data-collection strategy from Q1 compare to naively training on large open-source datasets?}
    \item \textbf{Q3 (Fig.~3): How does translation improve as we scale the diversity of source data?}
    \item \textbf{Q4 (Fig.~4): Do these trends hold on real robots (PiperX, WidowX, Franka, Piper)?}
\end{enumerate}

\subsection*{Simulation Environment}
We use a RoboCasa-based benchmark for cross-embodiment learning~\cite{robocasa2024}. Tasks: \emph{PnP Counter$\to$Sink}, \emph{PnP Sink$\to$Counter}, \emph{Turn On Sink Faucet}, and \emph{Flip Mug Upright}, where "PnP" stands for "Pick and Place". Scenes vary kitchen layout, textures, objects, and camera placement. Priors are generated with MimicGen~\cite{mandlekar2023mimicgen} for three embodiments (Kinova, Kinova3, UR5e) with Robotiq 2F-85/2F-140 grippers. In order to ensure that the simulation resets to the same state, we standardize the environment generation seed across experiments. Note that this is used mainly for evaluation consistency and not a strong assumption for data collection.

\subsection*{Real-World Environment}
We evaluate on three targets: \textbf{Franka}, \textbf{WidowX}, and \textbf{PiperX} (parallel-jaw). For each experiment, we collect $50$ demonstrations on the robot we want to transfer from. Then, we use our translational dataset collected as described below with $50$ demonstrations per axis, scene, and robot.

\subsection*{Experiment Setup}
\textbf{Policy and Inputs.}
We use a vision--language--action policy (\textbf{VLA}, $\pi_{0.5}$-style) $\pi_\theta(a_t \mid o_{1:t}, e, \ell)$ with an embodiment token $e$ and language task prompt $\ell$, where actions are represented in joint space. The policy is instantiated with $\pi_{0.5}$: a vision--language backbone that embeds the history of observations and language into tokens, followed by a flow-matching action expert that predicts a short-horizon sequence of joint actions. Concretely, we encode the third-person and wrist RGB images, proprioceptive state, embodiment token $e$, and text prompt $\ell$ into a sequence of tokens, append learned action tokens corresponding to a horizon-$H = 20$ sequence of joint position (or velocity/torque) commands, and process the resulting sequence with a small transformer. During training, we corrupt the ground-truth joint action sequence $a_{t:t+H-1}$ with Gaussian noise at a randomly sampled diffusion timestep and train the network to predict this noise, as in standard diffusion policies; at test time, we sample from the learned reverse process and execute only the first joint action $a_t$.

\textbf{Training.}
We start with the base $\pi_{0.5}$ VLA with pretrained weights \cite{intelligence2025pi05visionlanguageactionmodelopenworld}, then \emph{co-fine-tune} on the union of target few-shot demonstrations and a selected subset of source data, known as the "translation dataset" according to each study condition. Unless otherwise noted, we combine target and translation dataset at a $50{:}50$ ratio within each mini-batch. For the target-only and target upper-bound baselines, we finetune only with data from the robot we want to transfer from, or the robot we want to transfer to. Fine-tuning takes around $8$ hours on an NVIDIA A100 40GB GPU. Crucially, we do not modify the model architecture, loss functions, or optimization procedure; all experiments share the same training setup. Our study isolates the effect of \emph{data composition} during fine-tuning.

\textbf{Translation Data Collection.}
The translation dataset is constructed by systematically addressing each domain shift axis using the appropriate coverage and pairing strategy as determined by our ablation studies (Fig.~\ref{fig:main-coverage-sr}). 
\textit{In simulation}, for the \emph{viewpoint} axis, we leverage RoboCasa's procedural generation to sample diverse camera poses and intrinsics via changing the azimuth, elevation, and focal length. For the \emph{morphology} axis, we collect demonstrations across three robot platforms (Franka Emika Panda, Kinova3, UR5e) with different gripper types (Robotiq 2F-85/2F-140) using targeted coverage to span kinematic regimes and workspace geometries relevant to the target robot, with trajectory pairing obtained by filtering MimicGen-generated priors for high DTW similarity in end-effector and object keypoint trajectories. For the \emph{appearance} axis, we use RoboCasa's texture randomization to diversely vary kitchen materials, lighting, and object appearances.

\textit{In the real world}, for diversity in \emph{viewpoint} and \emph{morphology}, we vary the third-person camera and use the corresponding end-effectors. For \emph{appearance}, we augment the datasets with inpainted robots to the buffer using Dall-E 3. We keep the environment and state the same, only replacing the texture of the robot. Note that this process is a bit noisy, and sometimes textures that are not part of the robot get inpainted. However, in practice, these errors seem minimal as the model has qualitatively good performance.

\textbf{Data Budget.}
Unless otherwise stated, we use a fixed budget of \textbf{50 demonstrations per (robot, task)} combination for both translational data (or data that is included during co-finetuning to help the policy to understand cross-embodiment correspondences) and targets (data from another robot that we want to transfer). We report two references in every plot: \emph{Target-only} (few-shot baseline with no source data) and \emph{Target upper bound} (spending the same extra budget on the target robot). The results are compiled over random initializations $5$ in the real-world and random seeds $100$ in simulation. 

\textbf{Compositional data mixture (\emph{OXE+Translational}).}
For experiments that use \emph{OXE+Translational} (Fig.~\ref{fig:oxe_vs_targeted}, Fig.~\ref{fig:real_world}), the \emph{source} half of each co-fine-tuning batch is drawn as $60\%$ \emph{OXE} (unpaired) and $40\%$ \emph{Trajectory-Paired}. Within the \emph{OXE} pool, we reweight sampling to flatten histograms along viewpoint (camera azimuth/elevation bins) and morphology (gripper/kinematic class), avoiding over-representation of a few robots or camera setups. Across conditions, we fix the number of \emph{source} samples seen per epoch to isolate the effect of pairing versus volume.

\begin{figure}
    \centering
    \includegraphics[width=0.5\textwidth]{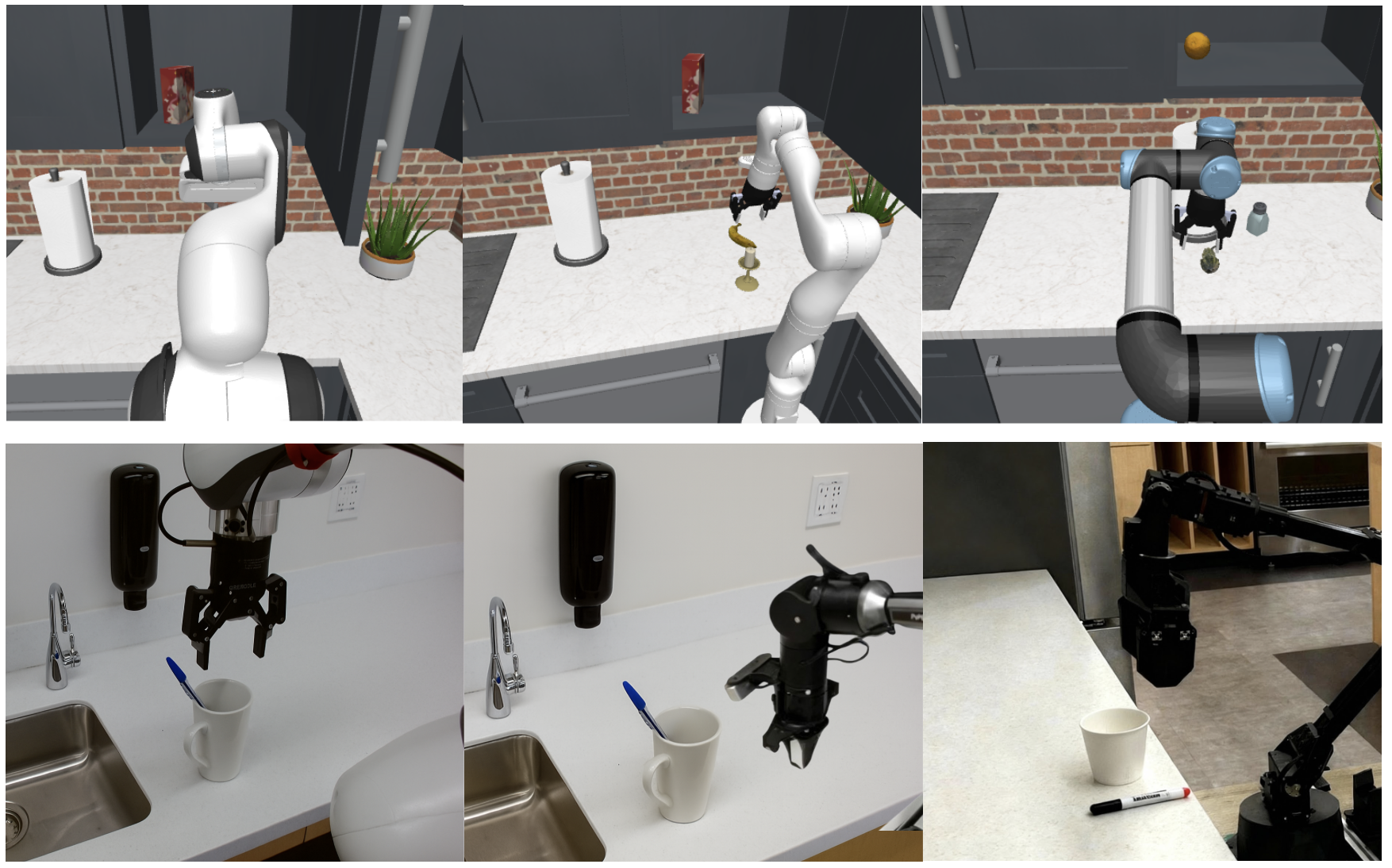}
    \caption{\textbf{Sim and Real-World Robots.} We train a \emph{cross-embodiment} policy to transfer the tasks of putting a pen in a cup and putting a book on a bookshelf to a new robot. We evaluate with the Franka Emika Panda, WidowX, and ARX Piper robot arms.}
    \label{fig:real-world}
\end{figure}

\textbf{Data Selection Methods.}
We instantiate four selection strategies that factor coverage and pairing:
\begin{itemize}
    \item \emph{Uniform (Naïve scaling):} uniformly sample source demos across embodiments.
    \item \emph{Targeted coverage:} select source demos to fill gaps in \emph{viewpoint} (camera pose/intrinsics) and \emph{morphology} (gripper/kinematics) relative to the target’s few-shot set; appearance diversity is secondary.
    \item \emph{Task-Paired:} for each target task instance, include source demos of the \emph{same task} (same objects/goals).
    \item \emph{Trajectory-Paired:} 
        \begin{itemize}
        \item \emph{Simulation:} For each trajectory, we select a counterpart by comparing proprioceptive features (end-effector pose, gripper state) and object keypoints. We define a task-specific \emph{event keypoint} $t^\star$ (e.g., first stable grasp) and align the approach segments (start $\rightarrow t^\star$) across trajectories using dynamic time warping (DTW) over a task-space feature $\phi_t=[x^{\mathrm{ee}}_t, R^{\mathrm{ee}}_t, g_t, \kappa_t]$, where $\kappa_t$ is an object-centric progress scalar (distance to the nearest task-relevant object keypoint in the object frame). Trajectories are downsampled to a fixed length (50 timesteps). The nearest neighbor under the DTW cost is chosen as the paired trajectory.
        \item \emph{Real World:} We collect demonstrations from different robots (e.g., a WidowX and a Franka) in the same task instance (same objects, layout, and goal) and then align them with DTW on $\phi_t$. In practice, the “identical scene” criterion means the same staged setup with matched object placements and camera viewpoints within normal lab tolerances; no special fiducials are required.
        \end{itemize}
\end{itemize}

\subsection{Under a fixed budget, what data collection strategy leads to the highest performance for each generalization axis?}
\begin{figure*}[t]
\centering
\includegraphics[width=\textwidth]{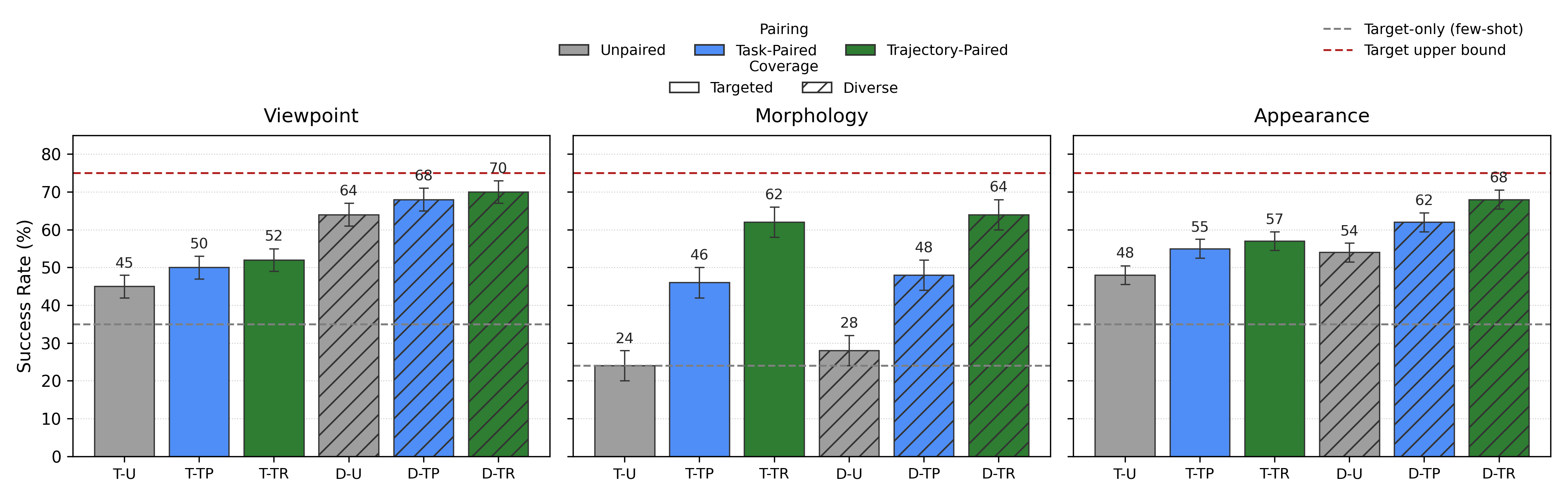}
\caption{\textbf{Main coverage plot (Success Rate).} Success Rate (\%) on the target robot across
Coverage$\times$Pairing for (a) Viewpoint, (b) Morphology, (c) Appearance. Error bars: 95\% CI.
Dashed lines show Target-only (few-shot) and Target upper bound (same extra budget on target).}
\label{fig:main-coverage-sr}
\end{figure*}

Figure~\ref{fig:main-coverage-sr} compares success across \emph{viewpoint}, \emph{morphology}, and \emph{appearance} under matched source budgets. In all settings, the target robot contributes only \textbf{50} demonstrations; we pre-train on sources and co-fine-tune on the union of this 50-shot target set and a selected subset of source data. Bars report means over 3--5 seeds for a fixed VLA trained with distinct \emph{coverage} (targeted vs.\ diverse) and \emph{pairing} (unpaired, task-paired, trajectory-paired) choices; dashed lines mark the \emph{Target-only} baseline and a \emph{Target upper bound} that spends the same extra budget directly on the target.

\textbf{Diversity helps for viewpoint and appearance. }\
For shifts dominated by perception---camera pose/intrinsics (\emph{viewpoint}) and scene/textures (\emph{appearance})---\emph{diverse} coverage outperforms \emph{targeted} at a fixed pairing level. Broad visual variation regularizes the encoder, reduces overfitting to scene- or camera-specific cues, and improves generalization even with weak alignment. In practice, sampling many camera poses, lighting regimes, and backgrounds produces larger gains than narrowly matching the target’s exact settings.

\textbf{Targeted coverage is more effective for morphology.}
When the distribution shift involves morphology, targeted selection provides a larger advantage than purely diverse sampling. As shown in Figure \ref{fig:main-coverage-sr}, performance along the morphology axis is similar for targeted and diverse selection ($62\%$ vs. $64\%$ for T-TR and D-TR), but the gap between paired and unpaired settings is substantially larger, averaging $23\%$. We hypothesize that this effect arises because different morphologies demand different object-manipulation strategies.

\subsection{How does our compositional data-collection strategy from Q1 compare to naively training on large open-source datasets?}
\label{subsec:q2}

\begin{figure*}[t]
\centering
\includegraphics[width=\textwidth]{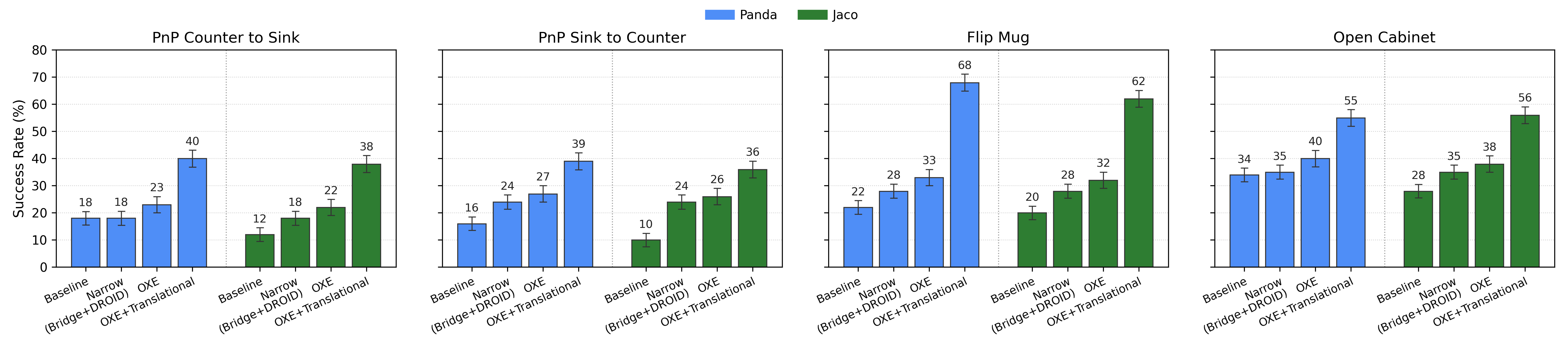}
\caption{\textbf{Comparison with large open-source datasets.} 
Success Rate (\%) on four RoboCasa tasks when training on (i) narrow two-robot data (\emph{Bridge+DROID}), (ii) large unpaired open-source datasets (\emph{OXE}), and (iii) our \emph{OXE+Translational} composition that reweights coverage and adds trajectory-level alignment. 
Results are shown for two target robots (\textbf{Panda}, \textbf{Jaco}); error bars denote 95\% CI.}
\label{fig:oxe_vs_targeted}
\end{figure*}

\begin{figure*}[t]
\centering
\includegraphics[width=\textwidth]{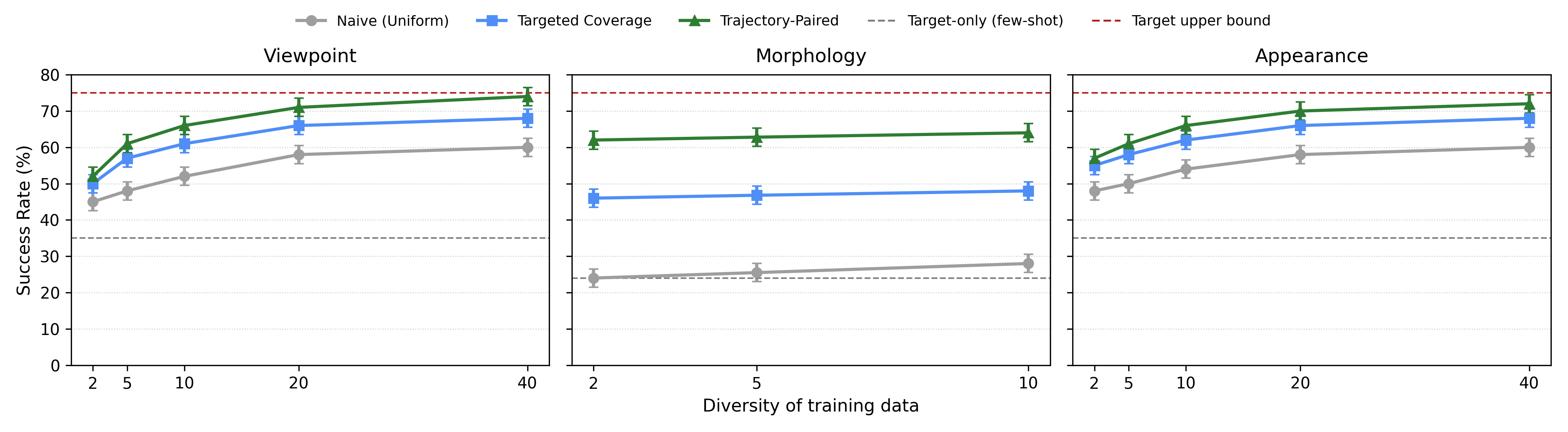}
\caption{\textbf{Effect of scaling diversity.}
Success Rate (\%) on the target robot as we increase the diversity of source data (distinct embodiments, viewpoints, and scenes). 
Curves compare \emph{Na\"ive/Uniform}, \emph{Targeted Coverage}, and \emph{Trajectory-Paired}. 
Dashed lines show the 50-shot \emph{Target-only} baseline and the \emph{Target upper bound}.}
\label{fig:scaling}
\end{figure*}

Figure~\ref{fig:oxe_vs_targeted} compares our compositional dataset design against training directly on large, unpaired open-source data. 
While large datasets such as OXE contain extensive demonstrations across robots and scenes, they are overwhelmingly \emph{unpaired} and emphasize scale over structure. 
We evaluate whether reweighting data coverage and introducing explicit translational pairs can outperform pure data volume.

\textbf{OXE is a strong baseline in simulation, but pairing leads to higher performance}
In Figure~\ref{fig:oxe_vs_targeted}, we compare our method against \emph{Baseline}, \emph{Narrow}, and \emph{OXE}. For \emph{Narrow} and \emph{OXE}, the translation dataset is replaced by the corresponding datasets (i.e., they are sampled as $50\%$ of each finetuning minibatch). While OXE provides a strong improvement over narrow paired data, our method consistently outperforms OXE across all tasks and both target robots, achieving an average of \textbf{$19\%$ higher success rate}. These gains are most pronounced on contact-rich tasks, indicating that explicit pairing and alignment are critical for fully leveraging large, unpaired datasets. This supports the view that while diversity at scale helps, explicit correspondences are a critical component to ensure that the morphology domain gap is crossed.

\textbf{Structured diversity is a valuable first step, but its gains are limited without explicit pairing.} The results also show that simply using a more diverse, structured data pool (\emph{OXE}) improves over a narrow, two-robot baseline (\emph{Bridge+DROID}), validating that broader coverage of cameras and scenes helps generalization. However, these gains from diversity alone plateau and are surpassed across tasks by our \emph{OXE+Translational} method. One contributing factor in simulation is that OXE’s real-world visual diversity does not perfectly align with simulated image statistics, limiting perception regularization.  Our work points to a data-scaling path that emphasizes composition—more pairing, broader coverage, and structured diversity—over simple aggregation.

Our compositional dataset differs from large unpaired sources along two key dimensions:
(1)~it \emph{balances coverage} across morphology and viewpoint axes to avoid over-representation of a few robots or camera setups, and 
(2)~it \emph{injects trajectory-level pairing} between related embodiments, encouraging policies to learn embodiment-invariant task representations. By ensuring that the domain gap is crossed for each axis, our method leads to significantly greater generalization than using unpaired data. In short, even under fixed budgets, scaling data \emph{structure} \emph{on top of} scaling data \emph{volume} delivers the best results: carefully composed cross-embodiment collections consistently add gains over strong unpaired baselines.

\subsection{How does translation improve as we scale the diversity of source data?}
\label{subsec:q3}

\begin{figure*}[t]
\centering
\includegraphics[width=\textwidth]{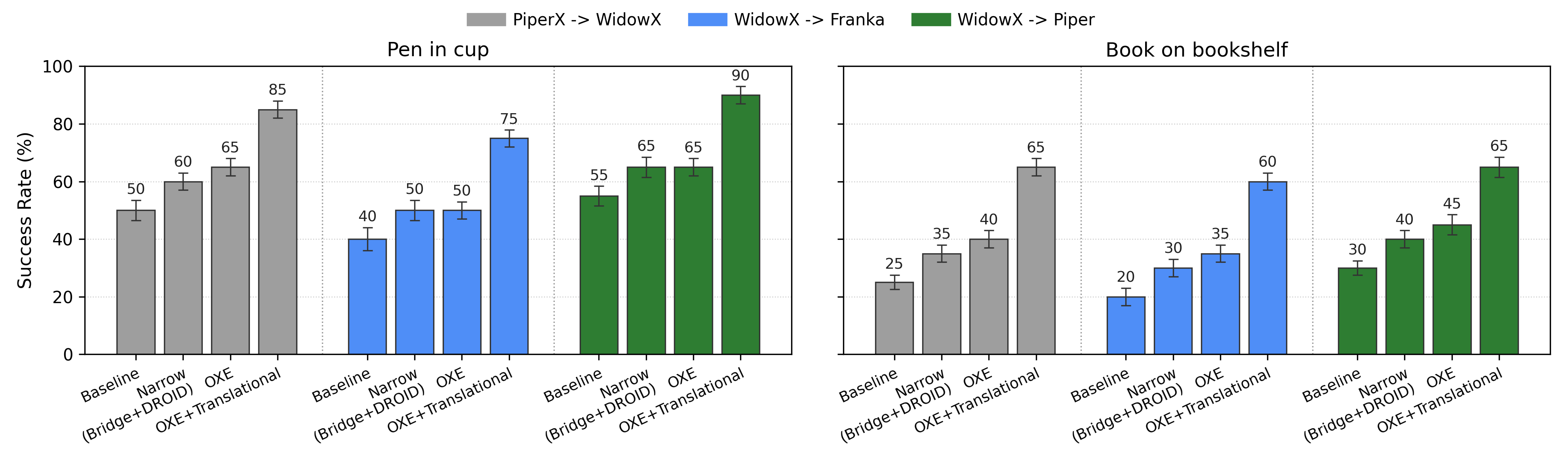}
\caption{\textbf{Real-robot transfer.} Success Rate (\%) on two tabletop tasks under three embodiment settings:
PiperX $\!\to$ WidowX, WidowX$\!\to$ Franka, and WidowX$\!\to$ Piper. 
Bars compare a few-shot target baseline, a narrow two-robot pool (\emph{Bridge+DROID}), a diversity-weighted open-source pool (\emph{OXE}), and our composed \emph{OXE+Translational}. Error bars denote 95\% CI.}
\label{fig:real_world}
\end{figure*}

\textbf{Viewpoint and appearance scale smoothly with diversity.}
For \emph{viewpoint} and \emph{appearance}, increasing diversity produces predictable, steady improvements across methods: adding more camera configurations and scene variations consistently lifts performance (an average increase of $17\%$). One plausible explanation is that broader coverage reduces distribution shift between source and target settings, making it more likely that the target’s conditions are represented during training. Notably, the success rate of \emph{trajectory pairing} maintains a consistent advantage even as diversity grows, by an average of $6\%$  over its less-paired counterpart—suggesting that aligning sequences across instances helps the model use this additional coverage more effectively rather than treating it as unrelated samples.

\textbf{Morphology saturates without pairing.}
On the \emph{morphology} axis, our results show that simply increasing the diversity of end-effectors does not enable generalization to new morphologies. Instead, strong cross-robot pairing is the dominant factor. This is evident in the data: increasing diversity (whether \emph{uniform} or \emph{targeted}) barely moves the performance curves ($42\%$ $\to$ $44\%$). Adding more arms or grippers without correspondences fails to resolve action-scale and kinematic mismatches; the policy sees different control distributions that cannot be reconciled by visual breadth alone. Temporal pairing, however, injects an object-frame alignment that effectively “translates” motion primitives across embodiments, converting cross-robot data into usable control supervision.

While data diversity solves the visual perception challenge—widening what the model can see and governing perception-level robustness—it is insufficient for action-level transfer. Instead, methods that encourage more correspondences between the two robots, such as task-paired or, more effectively, trajectory-paired data collection, are required. This "connectivity" provides the critical link that specifies how observations \emph{should} map to actions on a different robot.

\subsection{Do these trends hold on real robots (Franka, WidowX, Piper)?}
\label{subsec:q4}

\textbf{On real hardware, the benefits of translation data become more pronounced.}
Figure~\ref{fig:real_world} shows that the progression observed in simulation carries over to real robots: moving from a few-shot baseline to a narrow two-robot pool helps, reweighting large open-source data toward broader coverage helps more, and composing that pool with explicit cross-robot pairs delivers the largest improvements. Across embodiment settings that include both cross-platform transfer (PiperX$\!\to$ WidowX, WidowX$\!\to$ Franka) and a morphology change (WidowX$\!\to$ Piper), \emph{OXE+Translational} improves success by an average of $25\%$ over \emph{OXE}). This indicates that the benefits of cross-embodiment pairing are not simulator-specific. Instead, in the real-world, targeted data collection is even more necessary in order to cross the morphology gap between robots.

\begin{figure}
    \centering
    \includegraphics[width=0.5\textwidth]{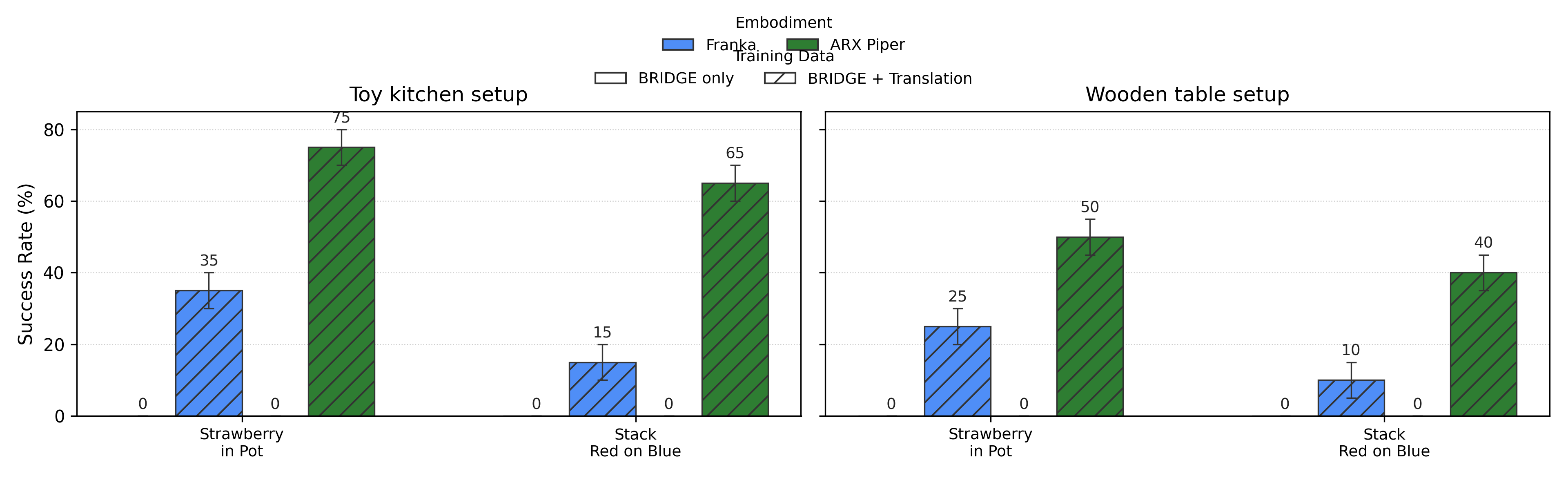}
    \caption{\textbf{BRIDGE dataset transfer.} We evaluate how well our method leads to transfer on two BRIDGE tasks across two environments each.}
    \label{fig:bridge_transfer}
\end{figure}

\textbf{Our translation dataset allows for transfer of tasks from existing open-source datasets.}
Figure~\ref{fig:bridge_transfer} depicts transfer results on two BRIDGE tasks, \emph{Strawberry in Pot} and \emph{Stack Red Block on Blue}. In order to test generalization to new scenes and avoid potential dataset leakage from the pretraining dataset of $\pi_{0.5}$, we evaluate each policy in two held-out real-world environments: a toy kitchen setup and a visually distinct wooden table setup. Notably, policies trained solely on BRIDGE fail to transfer, achieving $0\%$ success across all embodiments, tasks, and environments. However, when BRIDGE is augmented with paired translation data, cross-embodiment transfer performance is much higher.

For simpler manipulation behaviors that closely resemble motions present in the translation dataset—such as pick-and-place actions in \emph{Strawberry in Pot}—transfer performance is high, reaching up to $75\%$ success in the toy kitchen and $50\%$ on the wooden table. In contrast, more precision-sensitive behaviors such as block stacking exhibit lower, but still non-zero, success rates, improving from $0\%$ to as high as $65\%$ in the toy kitchen and $40\%$ on the wooden table. Performance consistently degrades in the wooden table setup, reflecting the added challenge of visual and geometric scene shifts. Overall, these results indicate that while large open-source datasets are not directly reusable across embodiments, a small amount of targeted translation data can unlock substantial transfer, particularly for manipulation skills that tolerate moderate spatial imprecision.

\textbf{There is substantial headroom in \emph{how} we scale robot data.}
These hardware results suggest that cross-embodiment transfer can be significantly improved by adding more structure to the data collection process. Rebalancing coverage across viewpoint and morphology already yields reliable gains over narrow pools, and introducing even modest amounts of cross-robot pairing correlates with further, repeatable improvements without collecting dramatically more demonstrations. In practice, this points to a concrete path for future datasets: allocate budget toward (i) diversity to cross the visual gap between robots and (ii) correspondences between embodiments. The consistent boosts on physical systems indicate that better-curated collections could unlock substantially stronger cross-robot generalization than current unpaired corpora afford.

\section{Conclusion}
\label{sec:conclusion}
By systematically varying viewpoint, morphology, and appearance under matched budgets, three design principles emerge. \emph{First}, breadth helps chiefly where variation is perceptual: increasing diversity in camera and scene coverage yields steady gains for viewpoint and appearance. \emph{Second}, morphology—the axis that changes how actions must be executed—benefits most from \emph{targeted} coverage. \emph{Third}, and most importantly, \emph{pairing} leads to large transfer benefits: introducing cross-robot correspondences consistently adds gains on top of strong unpaired baselines. Rather than accumulating more isolated islands of demonstrations, it is important to balance coverage along morphology and viewpoint, and invest a portion of budget in pairing that anchors examples in a common frame.

\section{Limitations}
\label{sec:limitations}
While our results show consistent gains from structured coverage and trajectory pairing, our study has several limitations. First, conclusions are drawn using a $\pi_{0.5}$-style VLA under a fixed few-shot budget; other architectures, larger budgets, or alternative training recipes could shift absolute performance and sometimes the relative gaps. Second, dataset shift remains: OXE and our scenes differ from simulation distributions and from other labs; our compositional strategy may require retuning to new object distributions, sensors, and embodiments. Finally, while our study drew conclusions from diverse simulations, our real-world experiments were conducted within two buildings. This means that it is unclear to what extent the translation dataset should scale to provide consistent general transfer between two real robots.

\section {Acknowledgements}
This work was supported in part by the Defense Advanced Research Projects Agency (DARPA) and the Toyota Research Institute (TRI). This material is also supported by the National Science Foundation under Grant No. 2125511.

\newpage

\bibliographystyle{plainnat}
\bibliography{references}

@inproceedings{khazatsky2024droid,
  title={{DROID: A Large-Scale In-the-Wild Robot Manipulation Dataset}},
  author={Khazatsky, Alexander and Pertsch, Karl and Nair, Suraj and Balakrishna, Ashwin and Dasari, Sudeep and Karamcheti, Siddharth and Nasiriany, Soroush and Srirama, Mohan Kumar and Chen, Lawrence Yunliang and Ellis, Kirsty and others},
  booktitle={Proceedings of Robotics: Science and Systems (RSS)},
  year={2024}
}

@inprocessings{Doshi24-crossformer,
  title={{Scaling Cross-Embodied Learning: One Policy for Manipulation, Navigation, Locomotion and Aviation}},
  author={Ria Doshi and Homer Walke and Oier Mees and Sudeep Dasari and Sergey Levine},
  booktitle={Conference on Robot Learning (CoRL)},
  year={2024}
}

@inproceedings{o2023open,
  title={{Open X-Embodiment: Robotic Learning Datasets and RT-X Models}},
  author={O'Neill, Abby and Rehman, Abdul and Gupta, Abhinav and Maddukuri, Abhiram and Gupta, Abhishek and Padalkar, Abhishek and Lee, Abraham and Pooley, Acorn and Gupta, Agrim and Mandlekar, Ajay and others},
  booktitle={International Conference on Robotics and Automation (ICRA)},
  year={2024},
}

@inproceedings{kim24openvla,
    title={{OpenVLA: An Open-Source Vision-Language-Action Model}},
    author={{Moo Jin} Kim and Karl Pertsch and Siddharth Karamcheti and Ted Xiao and Ashwin Balakrishna and Suraj Nair and Rafael Rafailov and Ethan Foster and Grace Lam and Pannag Sanketi and Quan Vuong and Thomas Kollar and Benjamin Burchfiel and Russ Tedrake and Dorsa Sadigh and Sergey Levine and Percy Liang and Chelsea Finn},
    booktitle={Conference on Robot Learning (CoRL)},
    year={2024}
}

@inproceedings{intelligence2025pi,
  title={{$\pi_ {0.5} $: a Vision-Language-Action Model with Open-World Generalization}},
  author={Intelligence, Physical and Black, Kevin and Brown, Noah and Darpinian, James and Dhabalia, Karan and Driess, Danny and Esmail, Adnan and Equi, Michael and Finn, Chelsea and Fusai, Niccolo and others},
  booktitle={Conference on Robot Learning (CoRL)},
  year={2025}
}

@inproceedings{szot2024multimodal,
  title={{From Multimodal LLMs to Generalist Embodied Agents: Methods and Lessons}},
  author={Szot, Andrew and Mazoure, Bogdan and Attia, Omar and Timofeev, Aleksei and Agrawal, Harsh and Hjelm, Devon and Gan, Zhe and Kira, Zsolt and Toshev, Alexander},
  booktitle={Conference on Computer Vision and Pattern Recognition (CVPR)},
  year={2025}
}

@inproceedings{dasari19robonet,
  author       = {Sudeep Dasari and
                  Frederik Ebert and
                  Stephen Tian and
                  Suraj Nair and
                  Bernadette Bucher and
                  Karl Schmeckpeper and
                  Siddharth Singh and
                  Sergey Levine and
                  Chelsea Finn},
  title        = {{RoboNet: Large-Scale Multi-Robot Learning}},
  booktitle={Conference on Robot Learning (CoRL)},
  year={2019}
}

@inproceedings{yang2024pushinglimitscrossembodimentlearning,
      title={{Pushing the Limits of Cross-Embodiment Learning for Manipulation and Navigation}}, 
      author={Jonathan Yang and Catherine Glossop and Arjun Bhorkar and Dhruv Shah and Quan Vuong and Chelsea Finn and Dorsa Sadigh and Sergey Levine},
      booktitle={Proceedings of Robotics: Science and Systems (RSS)},
      year={2024}
}

@inproceedings{hejna2024remixoptimizingdatamixtures,
      title={{Re-Mix: Optimizing Data Mixtures for Large Scale Imitation Learning}}, 
      author={Joey Hejna and Chethan Bhateja and Yichen Jian and Karl Pertsch and Dorsa Sadigh},
      booktitle = {Conference on Robot Learning (CoRL)},
      year      = {2024},
}

@inproceedings{brohan2022rt,
  title={{RT-1: Robotics Transformer for Real-World Control at Scale}},
  author={Brohan, Anthony and Brown, Noah and Carbajal, Justice and Chebotar, Yevgen and Dabis, Joseph and Finn, Chelsea and others},
  booktitle={Proceedings of Robotics: Science and Systems (RSS)},
  year={2023}
}

@inproceedings{brohan2023rt,
  title={{RT-2: Vision-Language-Action Models Transfer Web Knowledge to Robotic Control}},
  author={Brohan, Anthony and Brown, Noah and Carbajal, Justice and Chebotar, Yevgen and Chen, Xi and Choromanski, Krzysztof and others},
  booktitle =  {Conference on Robot Learning (CoRL)},
  year = 	 {2023},
}

@inproceedings{black2024pi_0,
  title={{$\pi_0$: A Vision-Language-Action Flow Model for General Robot Control}},
  author={Black, Kevin and Brown, Noah and Driess, Danny and Esmail, Adnan and Equi, Michael and Finn, Chelsea and Fusai, Niccolo and Groom, Lachy and Hausman, Karol and Ichter, Brian and others},
  booktitle={Proceedings of Robotics: Science and Systems (RSS)},
  year={2025}
}

@inproceedings{robocasa2024,
  title={{RoboCasa: Large-Scale Simulation of Everyday Tasks for Generalist Robots}},
  author={Soroush Nasiriany and Abhiram Maddukuri and Lance Zhang and Adeet Parikh and Aaron Lo and Abhishek Joshi and Ajay Mandlekar and Yuke Zhu},
  booktitle={Proceedings of Robotics: Science and Systems (RSS)},
  year={2024}
}

@misc{belkhale2024minivla,
      title={{MiniVLA: A Better VLA with a Smaller Footprint}}, 
      author={Suneel Belkhale and Dorsa Sadigh},
      url={https://github.com/Stanford-ILIAD/openvla-mini},
      year={2024},
}

@article{team2025gemini,
  title={{Gemini Robotics: Bringing AI into the Physical World}},
  author={Team, Gemini Robotics and Abeyruwan, Saminda and Ainslie, Joshua and Alayrac, Jean-Baptiste and Arenas, Montserrat Gonzalez and Armstrong, Travis and Balakrishna, Ashwin and Baruch, Robert and Bauza, Maria and Blokzijl, Michiel and others},
  journal={arXiv preprint arXiv:2503.20020},
  year={2025}
}

@inproceedings{
    yang2023polybot,
    title={{Polybot: Training One Policy Across Robots While Embracing Variability}},
    author={Jonathan Heewon Yang and Dorsa Sadigh and Chelsea Finn},
    booktitle={Conference on Robot Learning (CoRL)},
    year={2023},
}

@inproceedings{mandlekar2023mimicgen,
    title={{MimicGen: A Data Generation System for Scalable Robot Learning using Human Demonstrations}},
    author={Mandlekar, Ajay and Nasiriany, Soroush and Wen, Bowen and Akinola, Iretiayo and Narang, Yashraj and Fan, Linxi and Zhu, Yuke and Fox, Dieter},
    booktitle={Confernece on Robot Learning (CoRL)},
    year={2023}
}

@misc{intelligence2025pi05visionlanguageactionmodelopenworld,
      title={$\pi_{0.5}$: a Vision-Language-Action Model with Open-World Generalization}, 
      author={Physical Intelligence and Kevin Black and Noah Brown and James Darpinian and Karan Dhabalia and Danny Driess and Adnan Esmail and Michael Equi and Chelsea Finn and Niccolo Fusai and Manuel Y. Galliker and Dibya Ghosh and Lachy Groom and Karol Hausman and Brian Ichter and Szymon Jakubczak and Tim Jones and Liyiming Ke and Devin LeBlanc and Sergey Levine and Adrian Li-Bell and Mohith Mothukuri and Suraj Nair and Karl Pertsch and Allen Z. Ren and Lucy Xiaoyang Shi and Laura Smith and Jost Tobias Springenberg and Kyle Stachowicz and James Tanner and Quan Vuong and Homer Walke and Anna Walling and Haohuan Wang and Lili Yu and Ury Zhilinsky},
      year={2025},
      eprint={2504.16054},
      archivePrefix={arXiv},
      primaryClass={cs.LG},
      url={https://arxiv.org/abs/2504.16054}, 
}

@misc{wang2025unifiedvisionlanguageactionmodel,
      title={Unified Vision-Language-Action Model}, 
      author={Yuqi Wang and Xinghang Li and Wenxuan Wang and Junbo Zhang and Yingyan Li and Yuntao Chen and Xinlong Wang and Zhaoxiang Zhang},
      year={2025},
      eprint={2506.19850},
      archivePrefix={arXiv},
      primaryClass={cs.CV},
      url={https://arxiv.org/abs/2506.19850}, 
}

@misc{gao2024efficientdatacollectionrobotic,
      title={Efficient Data Collection for Robotic Manipulation via Compositional Generalization}, 
      author={Jensen Gao and Annie Xie and Ted Xiao and Chelsea Finn and Dorsa Sadigh},
      year={2024},
      eprint={2403.05110},
      archivePrefix={arXiv},
      primaryClass={cs.RO},
      url={https://arxiv.org/abs/2403.05110}, 
}

@INPROCEEDINGS{11127989,
  author={Kareer, Simar and Patel, Dhruv and Punamiya, Ryan and Mathur, Pranay and Cheng, Shuo and Wang, Chen and Hoffman, Judy and Xu, Danfei},
  booktitle={2025 IEEE International Conference on Robotics and Automation (ICRA)}, 
  title={EgoMimic: Scaling Imitation Learning via Egocentric Video}, 
  year={2025},
  volume={},
  number={},
  pages={13226-13233},
  doi={10.1109/ICRA55743.2025.11127989}}

@misc{xing2025shortcutlearninggeneralistrobot,
      title={Shortcut Learning in Generalist Robot Policies: The Role of Dataset Diversity and Fragmentation}, 
      author={Youguang Xing and Xu Luo and Junlin Xie and Lianli Gao and Hengtao Shen and Jingkuan Song},
      year={2025},
      eprint={2508.06426},
      archivePrefix={arXiv},
      primaryClass={cs.RO},
      url={https://arxiv.org/abs/2508.06426}, 
}

@misc{shi2025diversityneedscalablerobotic,
      title={Is Diversity All You Need for Scalable Robotic Manipulation?}, 
      author={Modi Shi and Li Chen and Jin Chen and Yuxiang Lu and Chiming Liu and Guanghui Ren and Ping Luo and Di Huang and Maoqing Yao and Hongyang Li},
      year={2025},
      eprint={2507.06219},
      archivePrefix={arXiv},
      primaryClass={cs.RO},
      url={https://arxiv.org/abs/2507.06219}, 
}

@misc{liu2025egozerorobotlearningsmart,
      title={EgoZero: Robot Learning from Smart Glasses}, 
      author={Vincent Liu and Ademi Adeniji and Haotian Zhan and Siddhant Haldar and Raunaq Bhirangi and Pieter Abbeel and Lerrel Pinto},
      year={2025},
      eprint={2505.20290},
      archivePrefix={arXiv},
      primaryClass={cs.RO},
      url={https://arxiv.org/abs/2505.20290}, 
}

@misc{doshi2024scalingcrossembodiedlearningpolicy,
      title={Scaling Cross-Embodied Learning: One Policy for Manipulation, Navigation, Locomotion and Aviation}, 
      author={Ria Doshi and Homer Walke and Oier Mees and Sudeep Dasari and Sergey Levine},
      year={2024},
      eprint={2408.11812},
      archivePrefix={arXiv},
      primaryClass={cs.RO},
      url={https://arxiv.org/abs/2408.11812}, 
}

@misc{lepert2025shadowleveragingsegmentationmasks,
      title={Shadow: Leveraging Segmentation Masks for Cross-Embodiment Policy Transfer}, 
      author={Marion Lepert and Ria Doshi and Jeannette Bohg},
      year={2025},
      eprint={2503.00774},
      archivePrefix={arXiv},
      primaryClass={cs.RO},
      url={https://arxiv.org/abs/2503.00774}, 
}

@misc{lepert2025masqueradelearninginthewildhuman,
      title={Masquerade: Learning from In-the-wild Human Videos using Data-Editing}, 
      author={Marion Lepert and Jiaying Fang and Jeannette Bohg},
      year={2025},
      eprint={2508.09976},
      archivePrefix={arXiv},
      primaryClass={cs.RO},
      url={https://arxiv.org/abs/2508.09976}, 
}

@misc{chen2024miragecrossembodimentzeroshotpolicy,
      title={Mirage: Cross-Embodiment Zero-Shot Policy Transfer with Cross-Painting}, 
      author={Lawrence Yunliang Chen and Kush Hari and Karthik Dharmarajan and Chenfeng Xu and Quan Vuong and Ken Goldberg},
      year={2024},
      eprint={2402.19249},
      archivePrefix={arXiv},
      primaryClass={cs.RO},
      url={https://arxiv.org/abs/2402.19249}, 
}

@misc{hu2025datascalinglawsimitation,
      title={Data Scaling Laws in Imitation Learning for Robotic Manipulation}, 
      author={Yingdong Hu and Fanqi Lin and Pingyue Sheng and Chuan Wen and Jiacheng You and Yang Gao},
      year={2025},
      eprint={2410.18647},
      archivePrefix={arXiv},
      primaryClass={cs.RO},
      url={https://arxiv.org/abs/2410.18647}, 
}

@misc{rayyan2025mvumiscalablemultiviewinterface,
      title={MV-UMI: A Scalable Multi-View Interface for Cross-Embodiment Learning}, 
      author={Omar Rayyan and John Abanes and Mahmoud Hafez and Anthony Tzes and Fares Abu-Dakka},
      year={2025},
      eprint={2509.18757},
      archivePrefix={arXiv},
      primaryClass={cs.RO},
      url={https://arxiv.org/abs/2509.18757}, 
}

@misc{liu2025immimiccrossdomainimitationhuman,
      title={ImMimic: Cross-Domain Imitation from Human Videos via Mapping and Interpolation}, 
      author={Yangcen Liu and Woo Chul Shin and Yunhai Han and Zhenyang Chen and Harish Ravichandar and Danfei Xu},
      year={2025},
      eprint={2509.10952},
      archivePrefix={arXiv},
      primaryClass={cs.RO},
      url={https://arxiv.org/abs/2509.10952}, 
}

@misc{pace2025xdiffusiontrainingdiffusionpolicies,
      title={X-Diffusion: Training Diffusion Policies on Cross-Embodiment Human Demonstrations}, 
      author={Maximus A. Pace and Prithwish Dan and Chuanruo Ning and Atiksh Bhardwaj and Audrey Du and Edward W. Duan and Wei-Chiu Ma and Kushal Kedia},
      year={2025},
      eprint={2511.04671},
      archivePrefix={arXiv},
      primaryClass={cs.RO},
      url={https://arxiv.org/abs/2511.04671}, 
}

@misc{cao2025gdreamgraphconditioneddiffusionretargeting,
      title={G-DReaM: Graph-conditioned Diffusion Retargeting across Multiple Embodiments}, 
      author={Zhefeng Cao and Ben Liu and Sen Li and Wei Zhang and Hua Chen},
      year={2025},
      eprint={2505.20857},
      archivePrefix={arXiv},
      primaryClass={cs.RO},
      url={https://arxiv.org/abs/2505.20857}, 
}

@misc{yan2024imitationnetunsupervisedhumantorobotmotion,
      title={ImitationNet: Unsupervised Human-to-Robot Motion Retargeting via Shared Latent Space}, 
      author={Yashuai Yan and Esteve Valls Mascaro and Dongheui Lee},
      year={2024},
      eprint={2309.05310},
      archivePrefix={arXiv},
      primaryClass={cs.RO},
      url={https://arxiv.org/abs/2309.05310}, 
}

@misc{choi2021selfsupervisedmotionretargetingsafety,
      title={Self-Supervised Motion Retargeting with Safety Guarantee}, 
      author={Sungjoon Choi and Min Jae Song and Hyemin Ahn and Joohyung Kim},
      year={2021},
      eprint={2103.06447},
      archivePrefix={arXiv},
      primaryClass={cs.RO},
      url={https://arxiv.org/abs/2103.06447}, 
}

@article{Aberman_2020,
   title={Skeleton-aware networks for deep motion retargeting},
   volume={39},
   ISSN={1557-7368},
   url={http://dx.doi.org/10.1145/3386569.3392462},
   DOI={10.1145/3386569.3392462},
   number={4},
   journal={ACM Transactions on Graphics},
   publisher={Association for Computing Machinery (ACM)},
   author={Aberman, Kfir and Li, Peizhuo and Lischinski, Dani and Sorkine-Hornung, Olga and Cohen-Or, Daniel and Chen, Baoquan},
   year={2020},
   month=aug }

@misc{bi2025hrdthumanmanipulationenhanced,
      title={H-RDT: Human Manipulation Enhanced Bimanual Robotic Manipulation}, 
      author={Hongzhe Bi and Lingxuan Wu and Tianwei Lin and Hengkai Tan and Zhizhong Su and Hang Su and Jun Zhu},
      year={2025},
      eprint={2507.23523},
      archivePrefix={arXiv},
      primaryClass={cs.RO},
      url={https://arxiv.org/abs/2507.23523}, 
}

@misc{yuan2025motiontranshumanvrdata,
      title={MotionTrans: Human VR Data Enable Motion-Level Learning for Robotic Manipulation Policies}, 
      author={Chengbo Yuan and Rui Zhou and Mengzhen Liu and Yingdong Hu and Shengjie Wang and Li Yi and Chuan Wen and Shanghang Zhang and Yang Gao},
      year={2025},
      eprint={2509.17759},
      archivePrefix={arXiv},
      primaryClass={cs.RO},
      url={https://arxiv.org/abs/2509.17759}, 
}

@misc{allu2025hrt1oneshothumantorobottrajectory,
      title={HRT1: One-Shot Human-to-Robot Trajectory Transfer for Mobile Manipulation}, 
      author={Sai Haneesh Allu and Jishnu Jaykumar P and Ninad Khargonkar and Tyler Summers and Jian Yao and Yu Xiang},
      year={2025},
      eprint={2510.21026},
      archivePrefix={arXiv},
      primaryClass={cs.RO},
      url={https://arxiv.org/abs/2510.21026}, 
}

@misc{nvidia2025gr00tn1openfoundation,
      title={GR00T N1: An Open Foundation Model for Generalist Humanoid Robots}, 
      author={NVIDIA and : and Johan Bjorck and Fernando Castañeda and Nikita Cherniadev and Xingye Da and Runyu Ding and Linxi "Jim" Fan and Yu Fang and Dieter Fox and Fengyuan Hu and Spencer Huang and Joel Jang and Zhenyu Jiang and Jan Kautz and Kaushil Kundalia and Lawrence Lao and Zhiqi Li and Zongyu Lin and Kevin Lin and Guilin Liu and Edith Llontop and Loic Magne and Ajay Mandlekar and Avnish Narayan and Soroush Nasiriany and Scott Reed and You Liang Tan and Guanzhi Wang and Zu Wang and Jing Wang and Qi Wang and Jiannan Xiang and Yuqi Xie and Yinzhen Xu and Zhenjia Xu and Seonghyeon Ye and Zhiding Yu and Ao Zhang and Hao Zhang and Yizhou Zhao and Ruijie Zheng and Yuke Zhu},
      year={2025},
      eprint={2503.14734},
      archivePrefix={arXiv},
      primaryClass={cs.RO},
      url={https://arxiv.org/abs/2503.14734}, 
}

@misc{chen2024roviaugrobotviewpointaugmentation,
      title={RoVi-Aug: Robot and Viewpoint Augmentation for Cross-Embodiment Robot Learning}, 
      author={Lawrence Yunliang Chen and Chenfeng Xu and Karthik Dharmarajan and Muhammad Zubair Irshad and Richard Cheng and Kurt Keutzer and Masayoshi Tomizuka and Quan Vuong and Ken Goldberg},
      year={2024},
      eprint={2409.03403},
      archivePrefix={arXiv},
      primaryClass={cs.RO},
      url={https://arxiv.org/abs/2409.03403}, 
}

\clearpage
\setcounter{page}{1}

\begin{figure*}[t]
\centering
\includegraphics[width=\textwidth]{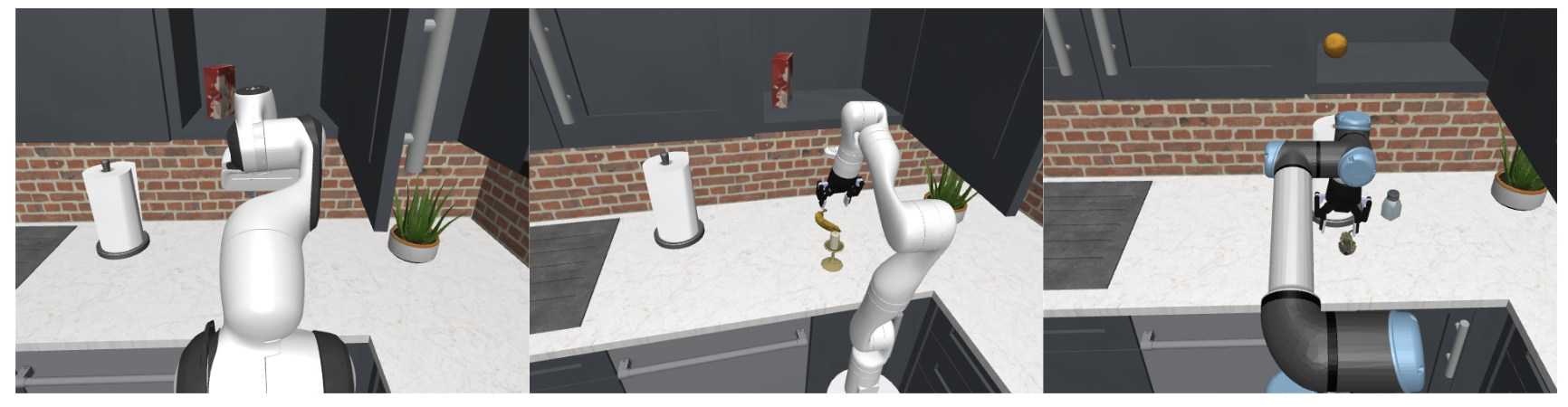}
\caption{\textbf{Simulation Robots.} Simulation robots with diverse camera angles and objects on a countertop scene.}
\label{fig:main-coverage-sr}
\end{figure*}

\begin{figure*}[t]
\centering
\includegraphics[width=\textwidth]{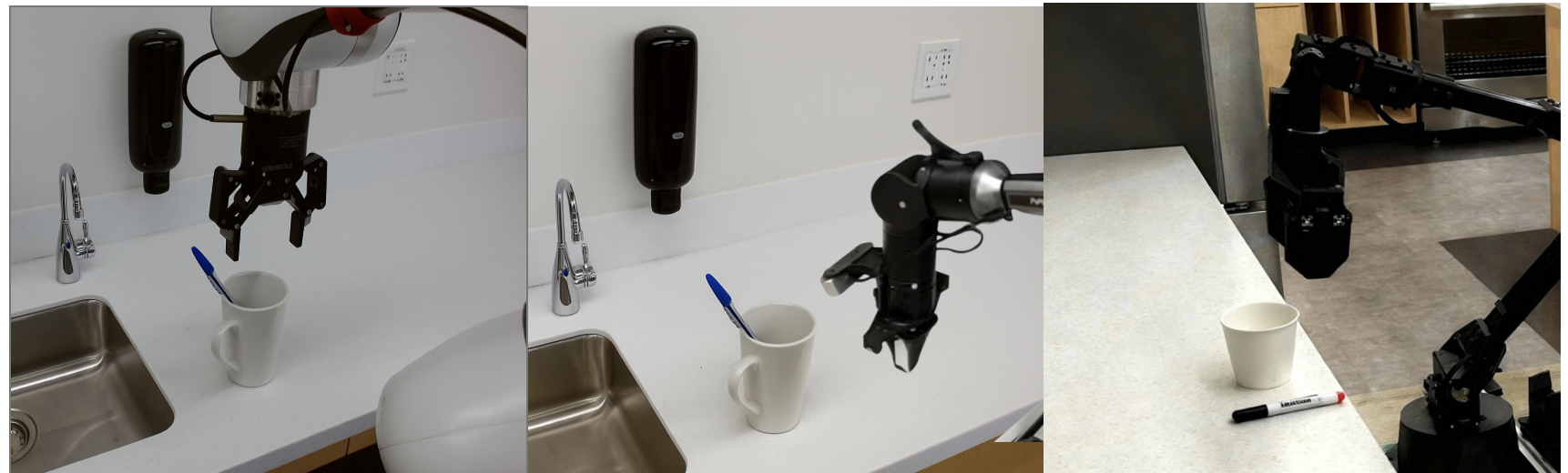}
\caption{\textbf{Real Robots.} Real robots with diverse camera angles and objects on a kitchen scene for the pen in cup task.}
\label{fig:main-coverage-sr}
\end{figure*}

\section{Website}
For further information, please visit 
\href{https://data-analogies.github.io/}{https://data-analogies.github.io/}

\section{Additional Simulation Details}
\subsection{Training Data}
We collect data on $10$ different scenes with $3$ different robots (UR5e, Kinova, Franka) on an Omrom base. Each (robot, scene) pair has $50$ different trajectories. For task and trajectory pairing, we choose $3$ more scenes to collect $50$ more trajectories each per robot in order to have enough data after filtering. For each state, we collect $3$ different observations at randomized camera angles in order to achieve diversity in the perspective axis.

\subsection*{Evaluation Tasks}
For our simulation evaluation, we focus on two primary pick-and-place tasks drawn from RoboCasa-X: \emph{PnP Counter$\to$Sink} and \emph{PnP Sink$\to$Counter}. We adapt these tasks slightly so that their object sets, layouts, and camera viewpoints are compatible with our real-world setups while preserving the core semantics of the original benchmarks. All tasks use a fixed third-person camera view of a single kitchen layout to facilitate consistent cross-embodiment comparisons.

\textbf{PnP Counter$\to$Sink.} As in RoboCasa-X, the robot must pick up an object from the countertop and place it anywhere inside the sink. In our variant, we limit the number of object categories to four and adjust the region from which the objects are initialized on the counter to avoid degenerate configurations (e.g., objects starting too close to the sink or outside the reachable workspace).

\textbf{PnP Sink$\to$Counter.} This task mirrors the reverse direction: the robot picks an object from the sink and places it on a plate positioned to the right of the sink. We use a single object category for this task to reduce ambiguity in the goal state and to align closely with our real-world evaluation protocol.

\section{Additional Real-World Details}

\subsection*{Robot Platforms}
Our real-world experiments utilize three distinct robotic platforms to evaluate cross-embodiment translation:

\begin{itemize}
    \item \textbf{Franka Emika Panda}: 7-DOF robotic arm with parallel-jaw gripper, workspace radius of approximately 85cm, and 6-axis force/torque sensor for contact sensing.
    \item \textbf{WidowX-250}: 6-DOF robotic arm manufactured by Trossen Robotics with parallel-jaw gripper, more compact workspace compared to Franka, and lower payload capacity (750g).
    \item \textbf{PiperX}: Compact desktop manipulator with 6-DOF and parallel-jaw end-effector, designed for tabletop manipulation tasks with high precision but limited reach.
\end{itemize}

\subsection*{Robot Controllers}
We use the DROID codebase~\cite{khazatsky2024droid} as our main wrapper for robot control, integrating platform-specific control interfaces for each robotic system:

\begin{itemize}
    \item \textbf{Franka Emika Panda}: Controlled via the DROID dataset which provides real-time Cartesian and joint-space control with an impedance controller.
    \item \textbf{WidowX-250}: Utilizes the official WidowX SDK from Trossen Robotics, providing direct servo control and position feedback through the onboard microcontroller.
    \item \textbf{PiperX}: Controlled through CAN bus communication, enabling precise joint-level control with real-time feedback from embedded encoders.
\end{itemize}

The DROID wrapper standardizes the control interface across all platforms, handling trajectory execution, safety monitoring, and data logging consistently regardless of the underlying hardware control method. 

\subsection*{Task Descriptions}
We focus on two primary manipulation tasks that require precise control and can be standardized across different robot platforms:

\begin{itemize}
    \item \textbf{Pen in Cup}: The robot must pick up a pen from a tabletop surface and place it vertically into a designated cup. This task requires precise grasping, trajectory planning to avoid collisions, and accurate placement with proper orientation.
    \item \textbf{Book on Bookshelf}: The robot must grasp a book lying flat on a table and place it vertically on a bookshelf slot. This task tests the ability to handle larger objects, manage different grasp poses, and execute placement with spatial precision.
\end{itemize}

\subsection{Scenes}
In order to ensure that the scenes collected in the training dataset do not lead to data leakage during evaluation, we collect data in $2$ different locations--one for training and one for evaluation. While both are indoor environments,  evaluation is done in a kitchen-like setting.

\subsection*{Data Collection Protocol}
For real-world data collection, we follow a standardized protocol to ensure consistency:

\begin{enumerate}
    \item \textbf{Environment Setup}: Each task is performed in a controlled tabletop environment with consistent lighting conditions. Objects are placed in predetermined positions marked by subtle visual cues to ensure repeatability.
    
    \item \textbf{Demonstration Collection}: Human demonstrators operate each robot using teleoperation to collect successful task demonstrations. We collect 50 demonstrations per robot for each task, ensuring diverse approach strategies while maintaining task success.
    
    \item \textbf{Cross-Robot Pairing}: For trajectory-paired data, we collect demonstrations of the same task instance (identical object placement and goal configuration) across different robots. These paired demonstrations are then temporally aligned using Dynamic Time Warping (DTW) on end-effector trajectories.
    
    \item \textbf{Viewpoint Variation}: Third-person cameras are positioned at multiple angles and heights to capture diverse viewpoints. Each robot setup includes both fixed third-person and wrist-mounted cameras.
    
    \item \textbf{Appearance Augmentation}: To increase visual diversity without requiring extensive real-world data collection, we employ DALL-E 3 for inpainting robot appearances while preserving scene context. This allows us to simulate different robot textures and colors within the same task episodes. The prompt we use is "Change the color of the robot with a different texture"
\end{enumerate}

\subsubsection{OXE Dataset Mixture}
The following table shows the data mixture for the OXE data: \\

\begin{table}[!htb]
    \centering
    \begin{tabular}{l|c}
    \toprule
    \textbf{Dataset} & \textbf{Split} \\
    \midrule
    Bridge & $12.5\%$ \\
    DROID & $12.5\%$ \\
    Fractal & $12.5\%$ \\
    Taco Play & $12.5\%$ \\
    Jaco Play & $12.5\%$ \\
    Roboturk & $12.5\%$ \\
    NYU Door Opening & $12.5\%$ \\
    Viola & $12.5\%$ \\
    \bottomrule
    \end{tabular}
    \caption{\textbf{Data Splits}}
    \label{table:manipeval}
\end{table}

\subsubsection{Data Loading}
We load all datasets through the \texttt{LeRobot} format, using the Open-X Embodiment (OXE) \texttt{repo\_id} as the primary entry point for each source. Each dataset is first converted into a standardized LeRobot schema that exposes RGB images, proprioception, actions, and task prompts under a common set of keys. We then apply dataset-specific \emph{repacking transforms} to map raw fields (e.g., joint positions, gripper states, multi-view camera streams) into this unified interface before feeding them to the Pi0.5 data pipeline.

For each embodiment, we reuse precomputed normalization statistics (means and variances) stored as dataset assets, ensuring consistent scaling across sources. The data pipeline composes three stages of transforms: (1) \emph{repack transforms} that remap dataset-specific keys into a common observation/action structure, (2) \emph{data transforms} that apply embodiment-specific preprocessing (e.g., converting absolute to delta joint actions, adapting ALOHA or DROID conventions to Pi0.5), and (3) \emph{model transforms} that resize images, tokenize the language prompt, and pad action/state sequences to match the model's action horizon. This design lets us train on heterogeneous LeRobot datasets without ad hoc sharding logic or manual trajectory slicing while keeping the alignment between observation and action spaces explicit.

\subsubsection{Model Training}
We fine-tune a Pi0.5-style VLA model on our compositional cross-embodiment dataset using the OpenPI training stack. Concretely, we begin from a Pi0.5 base checkpoint and perform LoRA-style low-rank adaptation on the vision–language backbone and action head while freezing the remaining weights. This yields a memory- and compute-efficient training setup that preserves the base model's broad visual-language competence while specializing a relatively small number of parameters for cross-embodiment translation. 

We co-fine-tune jointly on target and translation data, sampling batches with a 50{:}50 ratio between target episodes and source episodes drawn from the OXE+Translational mixture. Inside the OXE + Translational Mixture, we also weight the data by 50{:}50 when applicable (leading to a $25\%$, 25\%$, 50\%$ split between OXE, translational, and target data).

\begin{table}[!htb]
    \centering
    \caption{\textbf{Pi0.5 LoRA Training Hyperparameters.} Typical values used across our experiments.}
    \label{tab:pi05_hparams}
    \begin{tabular}{l|c}
    \toprule
    \textbf{Hyperparameter} & \textbf{Value} \\
    \midrule
    Base model & Pi0.5 VLA \\
    Fine-tuning strategy & LoRA \\
    Batch size (global) & 32 \\
    Optimizer & AdamW \\
    Learning rate schedule & Cosine decay with warmup \\
    Peak learning rate & $5\times10^{-5}$ \\
    Warmup steps & 10{,}000 \\
    Max token length & 180 (single-arm) \\
    Weight decay & $10^{-2}$ \\
    Target{:}source sampling ratio & 50{:}50 \\
    \bottomrule
    \end{tabular}
\end{table}

\section{Experiment Tables}

\subsection*{Table 1: Coverage vs. Pairing Results (Figure 1)}

\begin{table*}[t]
\centering
\caption{Success rates (\%) across coverage and pairing strategies for different domain shifts. Results show mean ± 95\% CI over 100 simulation trials.}
\label{tab:coverage_pairing}
\begin{tabular}{llccc}
\toprule
\textbf{Domain} & \textbf{Coverage} & \textbf{Unpaired} & \textbf{Task-Paired} & \textbf{Trajectory-Paired} \\
\midrule
\multirow{2}{*}{Viewpoint} & Targeted & $45.0 \pm 3.0$ & $50.0 \pm 3.0$ & $52.0 \pm 3.0$ \\
                           & Diverse  & $64.0 \pm 3.0$ & $68.0 \pm 3.0$ & $70.0 \pm 3.0$ \\
\midrule
\multirow{2}{*}{Morphology} & Targeted & $24.0 \pm 4.0$ & $46.0 \pm 4.0$ & $62.0 \pm 4.0$ \\
                            & Diverse  & $28.0 \pm 4.0$ & $48.0 \pm 4.0$ & $64.0 \pm 4.0$ \\
\midrule
\multirow{2}{*}{Appearance} & Targeted & $48.0 \pm 2.5$ & $55.0 \pm 2.5$ & $57.0 \pm 2.5$ \\
                           & Diverse  & $54.0 \pm 2.5$ & $62.0 \pm 2.5$ & $68.0 \pm 2.5$ \\
\bottomrule
\end{tabular}
\end{table*}

\textit{Key findings}: Target-only baseline: 35.0\%; Target upper bound: 75.0\%. For morphology, target-only performance is 24.0\%. Trajectory pairing consistently outperforms other strategies, with diverse coverage being most effective for viewpoint and appearance domains.

\subsection*{Table 2: Comparison with Open-Source Datasets (Figure 2)}

\begin{table*}[t]
\centering
\caption{Success rates (\%) comparing our compositional approach with large-scale open-source training. Results for Panda and Jaco target robots across four simulation tasks.}
\label{tab:oxe_comparison}
\begin{tabular}{llcccc}
\toprule
\textbf{Task} & \textbf{Robot} & \textbf{Baseline} & \textbf{Bridge+DROID} & \textbf{OXE} & \textbf{OXE+Translational} \\
\midrule
\multirow{2}{*}{PnP Counter→Sink} & Panda & $18.0 \pm 2.5$ & $18.0 \pm 2.6$ & $23.0 \pm 3.0$ & $\mathbf{40.0 \pm 3.1}$ \\
                                 & Jaco  & $12.0 \pm 2.5$ & $18.0 \pm 2.6$ & $22.0 \pm 3.0$ & $\mathbf{38.0 \pm 3.1}$ \\
\midrule
\multirow{2}{*}{PnP Sink→Counter} & Panda & $16.0 \pm 2.5$ & $24.0 \pm 2.6$ & $27.0 \pm 3.0$ & $\mathbf{39.0 \pm 3.1}$ \\
                                 & Jaco  & $10.0 \pm 2.5$ & $24.0 \pm 2.6$ & $26.0 \pm 3.0$ & $\mathbf{36.0 \pm 3.1}$ \\
\midrule
\multirow{2}{*}{Flip Mug} & Panda & $22.0 \pm 2.5$ & $28.0 \pm 2.6$ & $33.0 \pm 3.0$ & $\mathbf{68.0 \pm 3.1}$ \\
                         & Jaco  & $20.0 \pm 2.5$ & $28.0 \pm 2.6$ & $32.0 \pm 3.0$ & $\mathbf{62.0 \pm 3.1}$ \\
\midrule
\multirow{2}{*}{Open Cabinet} & Panda & $34.0 \pm 2.5$ & $35.0 \pm 2.6$ & $40.0 \pm 3.0$ & $\mathbf{55.0 \pm 3.1}$ \\
                             & Jaco  & $28.0 \pm 2.5$ & $35.0 \pm 2.6$ & $38.0 \pm 3.0$ & $\mathbf{56.0 \pm 3.1}$ \\
\bottomrule
\end{tabular}
\end{table*}

\textit{Key findings}: Our OXE+Translational approach consistently outperforms all baselines, with particularly large gains on complex manipulation tasks like Flip Mug (35-40 percentage point improvements over baseline).

\subsection*{Table 3: Real-World Transfer Results (Figure 4)}

\begin{table*}[t]
\centering
\caption{Real-world success rates (\%) across different robot transfer pairs. Results averaged over 5 seeds with 95\% confidence intervals.}
\label{tab:real_world}
\begin{tabular}{llcccc}
\toprule
\textbf{Task} & \textbf{Transfer} & \textbf{Baseline} & \textbf{Bridge+DROID} & \textbf{OXE} & \textbf{OXE+Translational} \\
\midrule
\multirow{3}{*}{Pen in Cup} & PiperX→WidowX  & $50 \pm 3.5$ & $60 \pm 3.0$ & $65 \pm 3.0$ & $\mathbf{85 \pm 3.0}$ \\
                           & WidowX→Franka  & $40 \pm 4.0$ & $50 \pm 3.5$ & $50 \pm 3.0$ & $\mathbf{75 \pm 3.0}$ \\
                           & WidowX→Piper   & $55 \pm 3.5$ & $65 \pm 3.5$ & $65 \pm 3.0$ & $\mathbf{90 \pm 3.0}$ \\
\midrule
\multirow{3}{*}{Book on Bookshelf} & PiperX→WidowX  & $25 \pm 2.5$ & $35 \pm 3.0$ & $40 \pm 3.0$ & $\mathbf{65 \pm 3.0}$ \\
                                  & WidowX→Franka  & $20 \pm 3.0$ & $30 \pm 3.0$ & $35 \pm 3.0$ & $\mathbf{60 \pm 3.0}$ \\
                                  & WidowX→Piper   & $30 \pm 2.5$ & $40 \pm 3.0$ & $45 \pm 3.5$ & $\mathbf{65 \pm 3.5}$ \\
\bottomrule
\end{tabular}
\end{table*}

\textit{Key findings}: Real-world results confirm simulation trends, with OXE+Translational achieving 25-35 percentage point improvements over baselines. The method is robust across different robot morphologies and task complexities.

\subsection*{Table 4: Scaling Analysis Summary (Figure 3)}

\begin{table*}[t]
\centering
\caption{Performance scaling with source diversity. Shows success rates at different numbers of source embodiments/viewpoints/scenes for each method.}
\label{tab:scaling}
\begin{tabular}{lllccccc}
\toprule
\textbf{Domain} & \textbf{Method} & \textbf{Baseline} & \textbf{2} & \textbf{5} & \textbf{10} & \textbf{20} & \textbf{40} \\
\midrule
\multirow{3}{*}{Viewpoint} & Naive (Uniform)   & \multirow{3}{*}{35.0} & $45.0$ & $48.0$ & $52.0$ & $58.0$ & $60.0$ \\
                           & Targeted Coverage &                       & $50.0$ & $57.0$ & $61.0$ & $66.0$ & $68.0$ \\
                           & Trajectory-Paired &                       & $52.0$ & $61.0$ & $66.0$ & $71.0$ & $74.0$ \\
\midrule
\multirow{3}{*}{Morphology} & Naive (Uniform)   & \multirow{3}{*}{24.0} & $24.0$ & $25.5$ & $28.0$ & --- & --- \\
                           & Targeted Coverage &                       & $46.0$ & $46.8$ & $48.0$ & --- & --- \\
                           & Trajectory-Paired &                       & $62.0$ & $62.8$ & $64.0$ & --- & --- \\
\midrule
\multirow{3}{*}{Appearance} & Naive (Uniform)   & \multirow{3}{*}{35.0} & $48.0$ & $50.0$ & $54.0$ & $58.0$ & $60.0$ \\
                           & Targeted Coverage &                       & $55.0$ & $58.0$ & $62.0$ & $66.0$ & $68.0$ \\
                           & Trajectory-Paired &                       & $57.0$ & $61.0$ & $66.0$ & $70.0$ & $72.0$ \\
\bottomrule
\end{tabular}
\end{table*}

\textit{Key findings}: Viewpoint and appearance benefits scale smoothly with diversity, while morphology shows saturation effects. Trajectory pairing maintains consistent advantages across all scaling levels. Target upper bound: 75.0\%.

\end{document}